\documentclass[sigconf, authorversion=true, nonacm=true]{acmart}

\usepackage{multirow}

\AtBeginDocument{%
  \providecommand\BibTeX{{%
    \normalfont B\kern-0.5em{\scshape i\kern-0.25em b}\kern-0.8em\TeX}}}

\begin{document}

\title{byteSteady: Fast Classification Using Byte-Level n-Gram Embeddings}

\author{Xiang Zhang}
\email{fancyzhx@gmail.com}
\affiliation{
  \institution{Element AI}
  \streetaddress{6650 Saint-Urbain}
  \city{Montreal}
  \state{Quebec}
  \country{Canada}
}

\author{Alexandre Drouin}
\email{alexandre.drouin@servicenow.com}
\affiliation{
  \institution{Element AI, ServiceNow}
  \streetaddress{6650 Saint-Urbain}
  \city{Montreal}
  \state{Quebec}
  \country{Canada}
}

\author{Raymond Li}
\email{raymond.li@servicenow.com}
\affiliation{
  \institution{Element AI, ServiceNow}
  \streetaddress{6650 Saint-Urbain}
  \city{Montreal}
  \state{Quebec}
  \country{Canada}
}


\begin{abstract}
This article introduces byteSteady -- a fast model for classification using byte-level \(n\)-gram embeddings. byteSteady assumes that each input comes as a sequence of bytes. A representation vector is produced using the averaged embedding vectors of byte-level \(n\)-grams, with a pre-defined set of \(n\). The hashing trick is used to reduce the number of embedding vectors. This input representation vector is then fed into a linear classifier. A straightforward application of byteSteady is text classification. We also apply byteSteady to one type of non-language data -- DNA sequences for gene classification. For both problems we achieved competitive classification results against strong baselines, suggesting that byteSteady can be applied to both language and non-language data. Furthermore, we find that simple compression using Huffman coding does not significantly impact the results, which offers an accuracy-speed trade-off previously unexplored in machine learning.
\end{abstract}




\maketitle

\section{Introduction}

Classification of sequential data such as text is a fundamental task in machine learning. Recently, tools such as fastText \cite{JGBM16} and Vowpal Wabbit \cite{WDLSA09} have shown that simple models can achieve state-of-the-art results for text classification. Moreover, it was shown that fastText could be trained at the \emph{character level} and still produce results competitive with a \emph{word-level} model~\cite{ZL17}.
This prompts us to explore an even lower level of data organization: \emph{byte level} and investigate whether training models at such a primal level allows their applicability beyond textual data.

In this article, we demonstrate a fastText-like model that is hard-coded to process input at the level of bytes, named \emph{byteSteady}, and show that it can achieve competitive classification results compared to word-level fastText models and byte-level convolutional networks. byteSteady assumes that inputs come as sequences of bytes. Like in fastText \cite{JGBM16}, a representation vector is produced using the averaged embedding vectors of byte-level \(n\)-grams, with a pre-defined set of \(n\). The hashing trick  \cite{WDLSA09} is used to reduce the number of embedding vectors. This input representation vector is then fed into a linear classifier to produce the desired class.

A byte-level sequence classification model has the potential to work beyond language data. In this article, we also apply byteSteady to gene classification for which the data comes as sequences of nucleotides. We show that byteSteady not only requires far less data pre-processing than previous standard models in bioinformatics, but also achieved better results for a large-scale gene classification data. Coincidentally, byte-level \(n\)-grams of nucleotides are also called \(k\)-mers in genomics~\cite{bonham2014alignment}.

When using a simple classifier such as logistic regression, much labor is usually required to design features under the assumption that raw signals are unlikely to be linearly separable. Conversely, deep learning models such as convolutional and recurrent networks are usually applied to sequential data at the raw signal level to extract hierarchical representation for classification. Results using byteSteady for text classification and gene classification have weakened this assumption and the methodological dichotomy, and paved the way for more exploration in using simple models for low-level inputs.

Byte-level inputs can also be compressed using standard data compression algorithms. In this article, we also show that for both text classification and gene classification tasks, the performance is not significantly impacted by compressing inputs using Huffman coding \cite{H52}. This provides a unique accuracy-speed trade-off that was not previously explored for machine learning.

\begin{figure*}[t]
  \centering
  \includegraphics[width=0.8\textwidth]{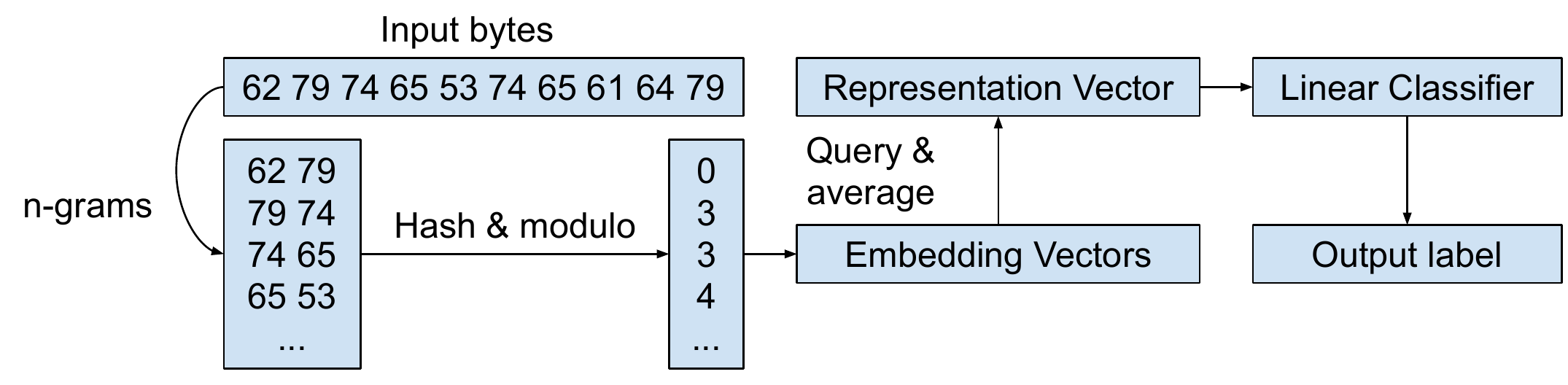}
  \caption{Illustration of byteSteady}
  \label{fig:model}
\end{figure*}

Prior work has explored byte-level text processing. For example, in \cite{GBVS16}, an LSTM-based \cite{HS97} sequence-to-sequence \cite{CMGBBSB14} \cite{SVL14} model is applied at byte level for a variety of natural language processing (NLP) tasks in 4 Romance languages. For NLP, the advantage of byte-level processing is that models can be immediately applied to any language regardless of whether there are too many entities at character or word levels. It alleviates the curse-of-dimensionality problem \cite{BDVJ03} with large vocabularies. Building on these advantages, we believe that this work is the first to explore the intersection of simple models (e.g., fastText-like) and byte-level processing, and demonstrate applicability outside the realm of language-based data.

\textbf{Contributions:} In a summary, this paper makes the following contributions: 1) demonstrate that byteSteady's byte-level encoding can achieve competitive performance for text classification; 2) extend its applicability beyond the scope of language data using gene classification as an example; 3) show that trivial speed-up and a new type of accuracy-speed tradeoff is possible using standard compression algorithms like Huffman coding. In addition, 4) a new large-scale gene classification dataset was built to promote further research in this domain.

\section{byteSteady Model}

As shown in figure \ref{fig:model}, byteSteady assumes that each sample consists of an input byte sequence and a output class label. The byte sequence is parsed into many byte-level \(n\)-grams using a pre-defined set of \(n\), which we call the \(n\)-gram set. Note that these \(n\)-grams are just sub-sequences of the input. For each \(n\)-gram, we compute a hash value, and use its modulo with respect to the pre-defined total number of embeddings as the index to query an embedding vector. This pre-defined number of embeddings is called the hash table size. Then, all of these embedding vectors are averaged together to produce the representation vector for the whole input byte sequence. This input representation vector is fed into a linear classifier to produce the desired class using the negative log-likelihood loss.

Except for the hard-coded byte-level sequence processing, byteSteady uses the same machine learning model as fastText \cite{JGBM16}. Combining the embedding table and the linear classifier, the entire model can be thought of as a 2-layer neural network without non-linearity. Assuming \(A\) is the embedding matrix and \(B\) is the linear classifier parameter matrix, the loss for each sample can be represented as
\begin{equation}
\label{eq:loss}
-y \log(\mathrm{softmax}(BAx)),
\end{equation}
in which \(y\) is the one-hot encoded label and \(x\) is the frequency-of-grams vector for the sample. \(A\) and \(B\) are the parameters to learn. It is well known that such a 2-layer neural network does not have more representational capacity than a linear model. In the case that the embedding dimension is larger than or equal to the number of classes, the capacity is the same.

For byte sequences such as texts and DNAs, \(x\) is usually sparse. This makes it possible to use online hashing, sparse matrix-vector multiplication and parallelization across samples for speeding up. Similar to fastText, we use the HogWILD! \cite{RRWF11} algorithm for fast learning on the CPU. The hashing trick \cite{WDLSA09} used in byteSteady has been shown to work well for word and character level text classification. We implemented Fowler–Noll–Vo (FNV) and CityHash as 2 hash function variants for byteSteady, and did not observe any difference in accuracy and speed.

Besides showing that such bag of tricks work well for text classification at the level of bytes, byte-level processing opens some other unique possibilities for classification task in general. The first is that byte-level processing can be applied to types of data other than text -- after all, all data in our computers are encoded as sequences of bytes. In this article, we explore gene classification as an example. Secondly, compression can be applied to byte-sequences. Since its outputs are also bytes, we could attempt to apply byteSteady on the shorter compressed byte sequence for classification. For both text classification and gene classification, we show that it is possible to find \(n\)-gram set configurations that work well for compressed data using Huffman coding \cite{H52}. This offers an accuracy-speed trade-off that is unexplored in machine learning before.

\section{Tasks}

In this section, we introduce the text classification and gene classification tasks with comparison between byteSteady and previous state-of-the-art models. Ablation studies for byteSteady on both of these tasks are presented in the next sections.

\subsection{Text Classification}

\begin{table*}[t]
  \caption{Text classification datasets and comparisons. For byteSteady, only one testing error is presented for each dataset in this article. The byteSteady model uses \(n\)-gram set \(\{4,8,12,16\}\) and a weight decay of 0.001, which was chosen after a hyper-parameter search using a development-validation split on each dataset. The embedding dimension is 16, and the hash table size is \(2^{24}=16,777,216\) The comparison models are the 5-gram word-level fastText model and large byte-level convolutional network (ConvNet) from \cite{ZL17}. Best result for each dataset is highlighted by a bold font.}
  \label{tab:text}
  \begin{center}
    \begin{tabular}{llcrrrrr}
      \hline
      \multicolumn{1}{c}{\textbf{Dataset}}  & \multicolumn{1}{c}{\textbf{Language}}  & \multicolumn{1}{c}{\textbf{Class}} & \multicolumn{1}{c}{\textbf{Train}} & \multicolumn{1}{c}{\textbf{Test}}& \multicolumn{1}{c}{\textbf{byteSteady}}& \multicolumn{1}{c}{\textbf{fastText}}& \multicolumn{1}{c}{\textbf{ConvNet}} \\ \hline
      Dianping & Chinese & 2 & 2,000K & 500K & \textbf{22.61\%} & 22.62\% & 23.17\% \\
      JD f. & Chinese & 5 & 3,000K & 250K & 50.55\% & 48.59\% & \textbf{48.10\%} \\
      JD b. & Chinese & 2 & 4,000K & 360K & 9.66\% & 9.93\% & \textbf{9.33\%} \\
      Rakuten f. & Japanese & 5 & 4,000K & 500K & \textbf{44.66\%} & 46.31\% & 45.10\% \\
      Rakuten b. & Japanese & 2 & 3,400K & 400K & \textbf{5.44\%} & 5.45\% & 5.93\% \\
      11st f. & Korean & 5 & 750K & 100K & 37.36\% & 38.67\% & \textbf{32.56\%} \\
      11st b. & Korean & 2 & 4,000K & 400K & 13.78\% & \textbf{13.23\%} & 13.30\% \\
      Amazon f. & English & 5 & 3,000K & 650K & 41.88\% & \textbf{40.02\%} & 42.21\% \\
      Amazon b. & English & 2 & 3,600K & 400K & 6.22\% & \textbf{5.41\%} & 6.52\% \\
      Ifeng & Chinese & 5 & 800K & 50K & 16.92\% & 16.95\% & \textbf{16.70\%} \\
      Chinanews & Chinese & 7 & 1,400K & 112K & 10.70\% & \textbf{9.24\%} & 10.62\% \\
      NYTimes & English & 7 & 1,400K & 105K & 14.81\% & \textbf{13.23\%} & 14.30\% \\
      Joint full & (Multiple) & 5 & 10,750K & 1,500K & 44.04\% & 43.29\% & \textbf{42.93\%} \\
      Joint binary & (Multiple) & 2 & 15,000K & 1,560K & \textbf{8.72\%} & 8.74\% & 8.79\% \\
      \hline
    \end{tabular}
  \end{center}
\end{table*}

The text classification datasets used in this article are the same as in \cite{ZL17}. In total, there are 14 large-scale datasets in 4 languages including Chinese, English, Japanese and Korean. There are 2 kinds of tasks in these datasets, which are sentiment analysis and topic classification. Moreover, 2 of the datasets are constructed by combining samples in all 4 languages to test the model's ability to handle different languages in the same fashion. Table \ref{tab:text} is a summarization.

In table \ref{tab:text}, the first 9 datasets are sentiment analysis datasets in either Chinese, English, Japanese or Korean, from either restaurant review or online shopping websites. The following 3 datasets are topic classification datasets in either Chinese or English, from online news websites. The last two are multi-lingual datasets by combining online shopping reviews from 4 languages.

It is worth noting that word-level or character-level models require significant pre-preprocessing for these languages, and the resulting vocabulary is large. It bears the risk of the curse-of-dimensionality \cite{BDVJ03}. Processing text at the level of bytes and using the hashing trick \cite{WDLSA09} alleviated these problems, and previous results \cite{GBVS16} \cite{ZL17} suggest that byte-level deep learning models could achieve competitive accuracy. However, as far as we know, we are the first to present results on byte-level text processing using simple models.

For hyper-parameter search, a development-validation split is constructed using a 90\%-10\% split on the training subset of each dataset. In the ablation study later, we will show that byteSteady results are insensitive to the hash table size and the embedding dimension as long as they are large enough. In this section, we use a hash table size of 16,777,216 (equals to \(2^{24}\)) and an embedding dimension of 16. Later ablation study will show that byteSteady results are sensitive to the \(n\)-gram set and the weight decay. For all datasets, the hyper-parameter search experiments suggest that the best \(n\)-gram set configuration is \(\{4,8,12,16\}\), and the best weight decay is 0.001.

Using these settings and the same comparison protocol as \cite{ZL17}, we present the testing errors in table \ref{tab:text} with comparisons to word-level 5-gram fastText models and byte-level one-hot encoded convolutional networks (ConvNets). These comparison models are the best of their kind for these datasets \cite{ZL17}, and represent strong baselines. The results in Table \ref{tab:text} suggest that byte-level simple models can achieve competitive results for text classification, sometimes surpassing previous state-of-the-arts. When byteSteady is not the state-of-the-art, it is not far from the best model.

\subsection{Gene Classification}

We now present an application of byteSteady to a kind of text-like non-language data -- DNA sequences for gene classification.
The use of $n$-gram representations of DNA sequences is increasingly prevalent in the field of genomics, as it allows to circumvent the computationally intensive process of multiple sequence alignment~\cite{bonham2014alignment}.
Such \textit{alignment-free} sequence analyses have been widely successful in problems such as DNA sequence assembly~\cite{boisvert2010ray, compeau2011apply}, phylogeny reconstruction and genome comparison~\cite{deraspe2017phenetic,leimeister2014kmacs,wen2014k}, genome-wide phenotype prediction~\cite{drouin2019interpretable,nguyen2018developing,davis2016antimicrobial,drouin2016predictive}, regulatory sequence prediction~\cite{ghandi2014enhanced,arvey2012sequence,sharmin2016heterogeneity}, and taxonomic profiling in metagenomics~\cite{raymond2016partial,vervier2016large,boisvert2012raymeta}.

\begin{table*}[t]
    \centering
    \caption{Results for the gene classification task. For each model, validation errors are shown for different \(n\)-gram sets. Arithmetic and set notations are used to provide shortened text. For example, \(2[1\text{-}8] = \{2,4,6,8,10,12,14,16\}\), and \(4^{[0\text{-}2]}=\{1,4,16\}\).}
    \begin{tabular}{lccccccccc}
        \toprule
        \(n\)-gram set & \(\{1\}\) & \(\{2\}\) & \(\{4\}\) & \(\{8\}\) & \(\{16\}\) & \(2 [1\text{-}8]\) & \(4 [1\text{-}4]\) & \(2^{[0\text{-}4]}\) & \(4^{[0\text{-}2]}\)\\
        \midrule
        Top-100K & \textbf{70.59\%} & \textbf{64.41\%} & \textbf{51.99\%} & 12.87\% & 43.50\% & 12.89\% & 12.85\% & 12.78\% & 32.68\% \\
        Top-1M & \textbf{70.59\%} & \textbf{64.41\%} & \textbf{51.99\%} & 12.87\% & 23.83\% & 13.88\% & 13.11\% & 12.49\% & 20.21\% \\
        byteSteady & 71.50\% & 64.70\% & 52.57\% & \textbf{11.57\%} & \textbf{6.81\%} & \textbf{3.79\%} & \textbf{4.12\%} & \textbf{7.03\%} & \textbf{13.76\%} \\
        \bottomrule
    \end{tabular}
    \label{tab:gene_results}
\end{table*}

The coupling of such representations with machine learning models has led to state-of-the-art results, but comes at the cost of gigantic feature spaces~\cite{drouin2019interpretable,vervier2016large}.
Factors that inflate the number of observed $n$-grams include the natural diversity of sequences (e.g., mutations) and random variations due to sequencing errors.
To tackle such huge feature spaces, some methods have relied on filtering $n$-grams as a preprocessing step~\cite{saeys2007review}, while others have turned to out-of-core data processing~\cite{drouin2016predictive,vervier2016large}.
The application of byteSteady to genomic data is therefore natural, as it can efficiently handle large $n$-gram feature spaces and can be directly applied to the DNA sequences in byte representation. In genomics, the \(n\)-grams of nucleotides are also called \(k\)-mers.

Our gene classification task consists of bacterial gene sequences in six high-level categories: antibiotic resistance, transporter, human homolog, essential gene, virulance factor, and drug target.
An accurate predictor for this task could improve automatic genome annotation and help detect important genes, such as those associated with the drug resistance and virulence of pathogenic bacteria.
We rely on a dataset extracted from the Pathosystems Resource Integration Center (PATRIC) database~\cite{davis2020patric,wattam2014patric}, which contains high-quality gene annotations for a large set of publicly available bacterial genomes.

The dataset contains a total of 5,111,616 DNA sequences, which are composed of nucleotides encoded as ASCII characters A, C, G, and T. The data is distributed evenly into $6$ classes with $851,936$ examples per class. We randomly select $90\%$ of the data for training and use the remaining $10\%$ for testing. We further partition the training set using the same proportions to produce a validation set for hyper-parameter search. Again, we use a hash table size of 16,777,216 (equals to \(2^{24}\)) and an embedding dimension of 16. The hyper-parameter search experiments suggest that the best \(n\)-gram set configuration is \(\{2,4,6,8,10,12,14,16\}\), and the best weight decay is 0.000001.

We show the benefits of using byteSteady in this context, by comparing to another linear classifier that requires manual $n$-gram feature selection due to computational constraints. This corresponds to the reality of practitioners in genomics, which must often resort to feature selection in order to apply standard machine learning algorithms to extremely large feature spaces~\cite{saeys2007review}.
We consider versions of this baseline that use the top 100,000 and 1,000,000 most frequent $n$-grams in the training data.

Our results in Table~\ref{tab:gene_results}, show that high prediction accuracies can be achieved, but only when considering large $n$-gram sizes (e.g., 16). This is only achievable for models like byteSteady, that have the ability to efficiently process the full set of features using the hashing trick \cite{WDLSA09}. For reference, the best byteSteady model, using \(n\)-gram set \(2[1-8] = \{2,4,6,8,10,12,14,16\}\) and weight decay \(10^{-6}\), achieves 3.73\% testing error. This is the only testing error for gene classification in this article.

\section{Ablation Study}

This section presents the ablation studies on 4 hyper-parameters in byteSteady - the \(n\)-gram set, the weight decay, the embedding dimension, and the hash table size. The general conclusion is that the results of byteSteady are highly sensitive to the \(n\)-gram set and the weight decay, but they do not improve much with increasing embedding dimension and the hash table size after a certain point.

For all of the ablation studies, we perform experiments on both text classification and gene classification. For the text classification task, we use the training subset of Dianping and make a 90\%-10\% development-validation split. All of the errors reported are on the validation datasets for both tasks.

\subsection{\(n\)-Gram Set and Weight Decay}

\begin{table*}[t]
  \caption{Ablation study on \(n\)-gram set and weight decay. All numbers are validation errors. The rows iterate through \(n\)-gram sets. Arithmetic and set notations are used to provide shortened text. For example, \(2[1\text{-}8] = \{2,4,6,8,10,12,14,16\}\), and \(4^{[0\text{-}2]}=\{1,4,16\}\). The columns are the weight decay parameters, which are differently chosen depending on the tasks. All of these experiments use an embedding dimension of 16 and a hash table size of \(2^{24}=16,777,216\). The best result for each task is highlighted using a bold font.}
  \label{tab:gram}
  \begin{center}
    \begin{tabular}{r|rrr|rrr}
      \hline
      & \multicolumn{3}{|c}{Text Classification}  & \multicolumn{3}{|c}{Gene Classification} \\
      & \(10^{-2}\) & \(10^{-3}\) & \(10^{-4}\) & \(10^{-5}\) & \(10^{-6}\) & \(10^{-7}\) \\ \hline
      \(\{1\}\) & 49.44\% & 41.54\% & 39.11\% & 71.28\% & 71.50\% & 71.38\% \\
      \(\{2\}\) & 40.48\% & 30.51\% & 28.19\% & 64.78\% & 64.70\% & 64.71\% \\
      \(\{4\}\) & 31.25\% & 27.11\% & 28.29\% & 52.60\% & 52.57\% & 52.55\% \\
      \(\{8\}\) & 26.83\% & 25.37\% & 26.96\% & 11.38\% & 11.57\% & 11.70\% \\
      \(\{16\}\) & 35.66\% & 32.77\% & 34.05\% & 5.38\% & 6.81\% & 7.93\% \\
      \([1-16]\) & 31.61\% & 26.16\% & 24.86\% & 4.27\% & 3.90\% & 4.17\% \\
      \(2[1-8]\) & 31.50\% & 25.50\% & 24.69\% & 3.86\% & \textbf{3.79\%} & 4.32\% \\
      \(4[1-4]\) & 28.97\% & \textbf{24.09\%} & 25.10\% & 3.90\% & 4.12\% & 4.76\% \\
      \(8[1-2]\) & 26.07\% & 24.73\% & 26.05\% & 4.79\% & 5.20\% & 5.89\% \\
      \(2^{[0-4]}\) & 33.99\% & 26.22\% & 26.94\% & 7.15\% & 7.03\% & 7.30\% \\
      \(4^{[0-2]}\) & 33.57\% & 27.44\% & 28.12\% & 15.97\% & 13.76\% & 14.26\% \\
      \(16^{[0-1]}\) & 45.97\% & 37.02\% & 38.66\% & 18.00\% & 22.23\% & 21.16\%\\
    \hline
    \end{tabular}
  \end{center}
\end{table*}

When using the byteSteady model for training, we need to provide an \(n\)-gram set for consideration. This is in contrast to some word-level models such as fastText \cite{JGBM16}, for which we only provide a single \(n\). In such a case, either only the gram set of \(\{n\}\) or \([1-n]\) is considered. Instead, we find that byteSteady is sensitive to the \(n\)-gram set, and the configuration \([1-n]\) does not perform the best.

Meanwhile, machine learning models are generally sensitive to the weight decay parameter. It is often used for the purpose of regularization, such that the gap between training and testing can become closer. This also applies to byteSteady. The best parameters depend on the task and the sample size.

Table \ref{tab:gram} details the results on the \(n\)-gram set and weight decay parameters. All of the numbers are validation errors. Variations of \(n\)-gram sets include the sets of a single \(n\), the sets of linearly increasing \(n\), and the sets of exponentially increasing \(n\). They all range up to \(n=16\). Due to the different in sample size, the best weight decay is different for each task. Text classification results are shown in \(\{10^{-2}, 10^{-3}, 10^{-4}\}\), whereas for gene classification they are in \(\{10^{-5}, 10^{-6}, 10^{-7}\}\). All of these experiments use an embedding dimension of 16 and a hash table size of \(2^{24}=16,777,216\).

There are a few conclusions from the results. The first is that longer \(n\)-grams give better results in the case of a single valued set. The second is that for both tasks, the best results come from a set of linearly increasing \(n\), albeit not \([1-16]\) which includes all \(n\)-grams in the range. We believe that this is because rich features like \([1-16]\) bear more risk of over-fitting.

The third conclusion is that the exponentially increasing \(n\)-gram set \(2^{[0-4]}=\{1,2,4,8,16\}\) does not perform significantly worse than the the linear sets. This offers an additional speed-accuracy trade-off, for which the computational complexity is \(\mathrm{O}(n)\) for the linear \(n\)-gram sets, and \(\mathrm{O}(\log(n))\) for the exponential alternatives. Whether the decrease in accuracy due to such reduction in computational complexity is acceptable should be dependent on the problem.

\subsection{Embedding Dimension and Hash Table Size}

Table \ref{tab:embed} details the validation errors for the ablation study on the embedding dimension for both text classification and gene classification. For these experiments, we use the \(n\)-gram set \(2^{[0-4]}=\{1,2,4,8,16\}\) and a hash table size of \(2^{24} = 16,777,216\). Different weight decay parameters are used for each task according to the previous ablation study. These results suggest that the embedding dimension does not significantly impact the model performance in terms of accuracy. However, it does directly impact the amount of memory needed to store the model parameters. Therefore, all of other experiments in this article always use an embedding dimension of 16 -- a moderate choice.

\begin{table}[h]
  \caption{Ablation study on embedding dimension. All of the numbers are validation errors. These experiments use the \(n\)-gram set \(2^{[0-4]}\) and a hash table size of \(2^{24} = 16,777,216\). For text classification, the weight decay used is 0.001. For gene classification, it is 0.000001.}
  \label{tab:embed}
  \begin{center}
    \begin{tabular}{lrrrrr}
      \hline
      &
      \multicolumn{1}{c}{\(4\)} &
      \multicolumn{1}{c}{\(8\)} &
      \multicolumn{1}{c}{\(16\)} &
      \multicolumn{1}{c}{\(32\)} & \multicolumn{1}{c}{\(64\)} \\ \hline
      Text & 26.13\% & 26.55\% & 26.22\% & 26.88\% & 26.16\% \\
      Gene & 7.14\% & 6.87\% & 7.30\% & 6.87\% & 7.27\% \\
    \hline
    \end{tabular}
  \end{center}
\end{table}

Table \ref{tab:hash} details the validation errors for the ablation study on the hash table size, for both text classification and gene classification. We choose the hash table size ranging from \(2^{16}=65,536\) to \(2^{26}=67,108,864\). These experiments use the \(n\)-gram set \(2^{[0-4]}=\{1,2,4,8,16\}\) and an embedding dimension of 16. Different weight decay parameters are used for each task according to the previous ablation study. From these results, we could conclude that the improvement can be observed when we increase the hash table size, but it becomes marginal after a certain point. As a result, all other experiments in this article use a moderately large hash table size \(2^{24}=16,777,216\).

\begin{table*}[t]
  \caption{Ablation study on hash table size. All of the numbers are validation errors. These experiments use the \(n\)-gram set \(2^{[0-4]}\) and an embedding dimension of 16. For text classification, the weight decay used is 0.001. For gene classification, it is 0.00001.}
  \label{tab:hash}
  \begin{center}
    \begin{tabular}{lrrrrrr}
      \hline
      &
      \multicolumn{1}{c}{\(2^{16}\)} &
      \multicolumn{1}{c}{\(2^{18}\)} &
      \multicolumn{1}{c}{\(2^{20}\)} &
      \multicolumn{1}{c}{\(2^{22}\)} & \multicolumn{1}{c}{\(2^{24}\)} & \multicolumn{1}{c}{\(2^{26}\)} \\ \hline
      Text & 27.11\% & 27.28\% & 27.14\% & 26.23\% & 26.22\% & 26.21\% \\
      Gene & 17.14\% & 9.61\% & 8.15\% & 7.12\% & 7.30\% & 7.01\% \\
    \hline
    \end{tabular}
  \end{center}
\end{table*}

Both ablation studies on the embedding dimension and the hash table size suggest that the byteSteady results are insensitive to these hyper-parameters when they are large enough. Therefore, future experiments will only focus on ablation studies for the \(n\)-gram set and the weight decay.

\section{Compression using Huffman Coding}

\begin{table*}[t]
  \caption{Huffman coding results. All numbers are validation errors. The rows iterate through different Huffman coding mechanism. The columns are the weight decay parameters, which are differently chosen depending on the tasks. All of these experiments use the \(n\)-gram set \(2^{[0-4]}=\{1,2,4,8,16\}\). The best result for each task in each coding mechanism is highlighted using a bold font. The best results using the same configuration without compression was 26.22\% for text classification, and 7.03\% for gene classification.}
  \label{tab:huffman}
  \begin{center}
    \begin{tabular}{r|rrr|rrr}
      \hline
      & \multicolumn{3}{|c}{Text Classification}  & \multicolumn{3}{|c}{Gene Classification} \\
      & \(10^{-2}\) & \(10^{-3}\) & \(10^{-4}\) & \(10^{-5}\) & \(10^{-6}\) & \(10^{-7}\) \\ \hline
      Bit 1 & 36.88\% & 27.72\% & \textbf{26.83\%} & \textbf{11.48\%} & 13.13\% & 14.85\% \\
      Bit 2 & 37.58\% & 29.03\% & 28.50\% & 16.48\% & 18.40\% & 20.45\% \\
      Bit 4 & 50.13\% & 35.66\% & 35.90\% & 24.28\% & 26.67\% & 29.03\% \\
      Bit 8 & 50.13\% & 47.93\% & 47.36\% & 31.89\% & 34.34\% & 36.44\% \\ \hline
      Byte 2 & 33.25\% & 27.09\% & \textbf{26.49\%} & \textbf{7.76\%} & 8.43\% & 8.67\% \\
      Byte 4 & 37.16\% & 28.78\% & 28.07\% & 10.47\% & 11.75\% & 13.10\% \\
      Byte 8 & 50.13\% & 38.70\% & 37.50\% & 15.37\% & 17.00\% & 18.56\% \\
    \hline
    \end{tabular}
  \end{center}
\end{table*}

This section presents a unique exploration enabled by byte-level data processing -- applying compression on the input byte sequences and presenting the resulting shorter sequences to byteSteady for classification. For text, previous character-level and word-level models could not apply because compression will render the character or word boundaries non-existent.

The compression algorithm we use here is Huffman coding \cite{H52}, using 2 variants for which the output are bits and bytes respectively. In both cases, we control the compression rate by limiting the byte length of symbols. We find that the model can perform well in low compression rates.

\subsection{Bit-Level and Byte-Level Huffman Coding}

\begin{table*}[t]
  \caption{Compression ratio for bit-level and byte-level Huffman coding. The compression ratio is defined as the faction of compressed size divided by the uncompressed size. The second row is the symbol length $m$ in the dictionary. For both text classification and gene classification tasks, we show the numbers for both the development and the validation subsets. For your reference the validation sets use the same compression dictionary obtained from the development sets. Symbol length 1 is ignored for byte-level Huffman coding because it offers no compression.}
  \label{tab:compression}
  \begin{center}
    \begin{tabular}{ll|rrrr|rrr}
      \hline
      & & \multicolumn{4}{c}{Bit level} & \multicolumn{3}{|c}{Byte level} \\
      & & \multicolumn{1}{c}{1} & \multicolumn{1}{c}{2} & \multicolumn{1}{c}{4} & \multicolumn{1}{c}{8} & \multicolumn{1}{|c}{2} & \multicolumn{1}{c}{4} & \multicolumn{1}{c}{8} \\ \hline
      \multirow{2}{*}{Text} & Development & 0.8615 & 0.6708 & 0.4894 & 0.3534 & 0.7133 & 0.5137 & 0.3644 \\
      & Validation & 0.8616 & 0.6707 & 0.4890 & 0.3454 & 0.7131 & 0.5155 & 0.4064 \\ \hline
      \multirow{2}{*}{Gene}  & Development & 0.4041 & 0.3161 & 0.2796 & 0.2606 & 0.5008 & 0.2527 & 0.2531 \\
      & Validation & 0.4041 & 0.3161 & 0.2796 & 0.2606 & 0.5008 & 0.2527 & 0.2531 \\
    \hline
    \end{tabular}
  \end{center}
\end{table*}

Huffman coding \cite{H52} works by giving shorter codes for higher frequency symbols, using a frequency-sorted tree for which the leaf nodes are symbols. In this article, symbols are defined as a byte sub-sequences of length \(m\). When the tree is binary, Huffman coding outputs binary codes (bits) one at a time. We call this variant the bit-level Huffman coding. On the other hand, if the tree is 256-ary, the codes can be generated one byte at a time. We name this variant the byte-level Huffman coding.

The symbol length \(m\) can exponentially impact the size of the frequency table. Naturally, larger \(m\) ensures longer symbols can be considered for shorter codes, and results in better compression rate. Therefore, \(m\) can be used to control the compression level of data, and offers a speed-accuracy trade-off that was not explored in machine learning before. Note that using \(m=1\) for byte-level Huffman coding will not compress because the symbols and the codes have the same length.

Table \ref{tab:compression} illustrates this property by presenting the compression ratio for the development and validation datasets of each task. The difference between development and validation is small in spite of using the development dictionary to compress the validation data. These ratio numbers can be directly translated to the reduction in training time when using compressed input.

The results for Huffman coding experiments are presented in table \ref{tab:huffman}. All of the experiments use the \(n\)-gram set \(2^{[0-4]} = \{1,2,4,8,16\}\). According to previous results, the best uncompressed model for this configuration achieves 26.22\% for text classification, and 7.01\% for gene classification. Given the best compressed result of 26.49\% and 7.76\% using the byte-level Huffman coding with symbol length \(m=2\), we know that a low level of compression does not affect the results significantly.

\begin{table*}[t]
  \caption{Ablation study on \(n\)-gram set and weight decay using a bit-level Huffman coding of symbol length 1. All numbers are validation errors. The rows iterate through \(n\)-gram sets. Arithmetic and set notations are used to provide shortened text. For example, \(2[1\text{-}8] = \{2,4,6,8,10,12,14,16\}\), and \(4^{[0\text{-}2]}=\{1,4,16\}\). The columns are the weight decay parameters, which are differently chosen depending on the tasks. The best result for each task is highlighted using a bold font.}
  \label{tab:huffman-gram}
  \begin{center}
    \begin{tabular}{r|rrr|rrr}
      \hline
      & \multicolumn{3}{|c}{Text Classification}  & \multicolumn{3}{|c}{Gene Classification} \\
      & \(10^{-2}\) & \(10^{-3}\) & \(10^{-4}\) & \(10^{-5}\) & \(10^{-6}\) & \(10^{-7}\) \\ \hline
      \(\{1\}\) & 50.13\% & 43.63\% & 42.04\% & 64.33\% & 64.33\% & 64.33\% \\
      \(\{2\}\) & 37.33\% & 31.35\% & 30.59\% & 51.65\% & 51.53\% & 51.50\% \\
      \(\{4\}\) & 29.79\% & 26.93\% & 28.99\% & 10.21\% & 11.69\% & 13.10\% \\
      \(\{8\}\) & 29.50\% & 28.54\% & 29.90\% & 22.05\% & 24.89\% & 26.72\% \\
      \(\{16\}\) & 50.12\% & 48.50\% & 47.54\% & 33.35\% & 35.65\% & 37.14\% \\
      \([1-16]\) & 38.76\% & 27.30\% & \textbf{26.12\%} & 12.62\% & 14.04\% & 16.30\% \\
      \(2[1-8]\) & 34.42\% & 27.09\% & 26.38\% & 13.28\% & 15.28\% & 17.60\% \\
      \(4[1-4]\) & 30.54\% & 26.06\% & 26.80\% & 12.87\% & 15.57\% & 18.29\% \\
      \(8[1-2]\) & 34.81\% & 29.22\% & 30.54\% & 24.80\% & 27.69\% & 30.02\% \\
      \(2^{[0-4]}\) & 36.88\% & 27.72\% & 26.83\% & 11.48\% & 13.13\% & 14.85\% \\
      \(4^{[0-2]}\) & 35.96\% & 27.55\% & 27.36\% & \textbf{11.15\%} & 12.84\% & 14.37\% \\
      \(16^{[0-1]}\) & 50.13\% & 43.61\% & 42.24\% & 28.97\% & 30.19\% & 30.78\% \\
    \hline
    \end{tabular}
  \end{center}
\end{table*}

On the other hand, for both bit-level and byte-level Huffman coding, we could observe that the results become worse as more aggressive compression is used. This offers a unique speed-accuracy trade-off based on input compression, which is an idea previously unexplored in machine learning. This could be useful for devices in a constrained computation and network environment.

If we compare between bit-level Huffman coding of symbol length 1 with byte-level of length 2 -- the lowest compression rates respectively -- byte-level Huffman coding gives better results for all the tasks. This is because generating byte-level codes preserves byte boundaries -- that is, byte boundaries in the compressed data were still byte boundaries in the original data. Bit-level Huffman coding does not preserve boundaries, which may be challenging for byteSteady to perform well.

\subsection{Ablation Study on \(n\)-Gram Set and Weight Decay}

Table \ref{tab:huffman-gram} details the results for an ablation study on \(n\)-gram set and weight decay, using the bit-level Huffman coding of symbol length 1. Similar to previous ablation study with uncompressed input, a composite \(n\)-gram set usually works better than a single choice of \(n\).

However, unlike in the uncompressed ablation study, the results for a single choice of \(n\) does not necessarily improve as \(n\) increases. Meanwhile, linearly increasing \(n\)-gram set also does not necessarily perform the best. These results suggest that hyper-parameter search should conduct differently when compression is used to speed up byteSteady at the input level.

\section{Conclusion}

In this article, we introduce byteSteady -- a fast model for classification using byte-level \(n\)-gram embeddings. The model produces a representation vector using the averaged embedding vectors of byte-level \(n\)-grams, and feeds it into a linear classifier for classification. The model is the same as fastText \cite{JGBM16}, except for the hard-coded byte-level input processing mechanism.

byteSteady can be applied to language and non-language data. In this article, we show experiments in text classification and gene classification. Competitive results against strong baselines are achieved. Since byteSteady reads input at the level of bytes, compression can be applied to the input for speeding up. We show that a low-level of compression using Huffman coding does not significantly impact the results, and provide a new speed-accuracy trade-off previously unexplored in machine learning.

In the future, we hope to extend byteSteady to unsupervised embedding learning for byte-level \(n\)-grams, and use it for more kinds of data and problems.

\bibliographystyle{ACM-Reference-Format}
\bibliography{article}

\end{document}